\PassOptionsToPackage{table}{xcolor}

\documentclass[sigconf,nonacm,balance=false]{acmart}
\setcopyright{none}
\makeatletter
\let\hyxmp@parse@acmart\relax
\makeatother
\usepackage{graphicx}
\usepackage{booktabs}
\usepackage{array}
\usepackage{multirow}
\usepackage{amsmath}

\usepackage{amssymb}
\usepackage{mathtools}
\usepackage{xspace}
\usepackage{tikz}
\usetikzlibrary{positioning,arrows.meta,calc,fit,shapes.geometric,backgrounds}
\usepackage{pgfplots}
\pgfplotsset{compat=1.18}
\usepackage{algorithm}
\usepackage{algpseudocode}

\makeatletter
\let\laplace\@undefined
\makeatother

\usepackage{amsmath,amsfonts,bm}

\def\eqref#1{equation~\ref{#1}}

\def\1{\bm{1}}

\def\vtheta{{\bm{\theta}}}

\def\ve{{\bm{e}}}

\def\vm{{\bm{m}}}

\def\vr{{\bm{r}}}

\def\vw{{\bm{w}}}

\def\vy{{\bm{y}}}

\def\mA{{\bm{A}}}

\def\mC{{\bm{C}}}

\def\mH{{\bm{H}}}

\def\mK{{\bm{K}}}

\def\mM{{\bm{M}}}

\def\mP{{\bm{P}}}
\def\mQ{{\bm{Q}}}
\def\mR{{\bm{R}}}

\def\mU{{\bm{U}}}
\def\mV{{\bm{V}}}
\def\mW{{\bm{W}}}

\def\mZ{{\bm{Z}}}

\DeclareMathAlphabet{\mathsfit}{\encodingdefault}{\sfdefault}{m}{sl}
\SetMathAlphabet{\mathsfit}{bold}{\encodingdefault}{\sfdefault}{bx}{n}

\def\sR{{\mathbb{R}}}

\newcommand{\laplace}{\mathrm{Laplace}} 

\newcommand{\E}{\mathbb{E}}

\newcommand{\softmax}{\mathrm{softmax}}

\usepackage[capitalise,noabbrev]{cleveref}

\newcommand{\ours}{FToken\xspace}
\newcommand{\clip}{\operatorname{clip}}

\newtheorem{proposition}{Proposition}

\definecolor{tabfirst}{rgb}{1,0.7,0.7}
\definecolor{tabsecond}{rgb}{1,0.85,0.7}
\definecolor{tabthird}{rgb}{1,1,0.7}
\newcommand{\first}[1]{\cellcolor{tabfirst}\textbf{#1}}
\newcommand{\second}[1]{\cellcolor{tabsecond}#1}
\newcommand{\third}[1]{\cellcolor{tabthird}#1}
\newcommand{\s}[1]{\textcolor{gray}{\scriptsize{$\pm #1$}}}

\graphicspath{{images/}}

\definecolor{dsAlpha158}{rgb}{0.84, 0.33, 0.30}  
\definecolor{dsAlpha360}{rgb}{0.27, 0.46, 0.71}  
\definecolor{dsAshare}  {rgb}{0.88, 0.55, 0.16}  

\pgfplotsset{
  /pgfplots/ft plot/.style={
    grid=both,
    grid style={line width=0.2pt, draw=gray!22},
    major grid style={line width=0.2pt, draw=gray!30},
    tick label style={font=\scriptsize},
    label style={font=\scriptsize},
    title style={font=\scriptsize, yshift=-2pt},
    every axis plot/.append style={line width=0.6pt, mark size=1.2pt},
    axis line style={line width=0.4pt, draw=gray!55!black},
    tick style={line width=0.3pt, color=gray!55!black},
  },
}

\tikzset{
  ft cell/.style={
    draw, line width=0.5pt, rounded corners=2pt,
    minimum height=11mm,
    align=center, inner xsep=5pt, inner ysep=4pt,
  },
  ft ours/.style={
    ft cell,
    draw=tabfirst!75!black, fill=tabfirst!45,
    line width=0.6pt,
  },
  ft io/.style={
    ft cell,
    draw=gray!60, fill=gray!12,
    minimum width=20mm,
  },
  ft arrow/.style={
    -{Latex[length=1.8mm,width=1.5mm]},
    line width=0.55pt, gray!50!black,
  },
  ft guide/.style={
    dashed, gray!60, line width=0.4pt,
  },
}

\title[FToken: Factors as Tokens for Dual-Axis Cross-Sectional Return Prediction]{FToken: Factors as Tokens for\texorpdfstring{\\}{ }Dual-Axis Cross-Sectional Return Prediction}

\author{Juntang Wang}
\affiliation{\institution{Harvard University}\city{Cambridge}\country{USA}}
\email{juntangwang@g.harvard.edu}

\author{Yihan Wang}
\affiliation{\institution{Duke Kunshan University}\city{Kunshan}\country{China}}
\email{yihan.wang@dukekunshan.edu.cn}

\author{Sheng'en Li}
\affiliation{\institution{Duke Kunshan University}\city{Kunshan}\country{China}}
\email{shengen.li@dukekunshan.edu.cn}

\author{Dongmian Zou}
\authornote{Corresponding authors.}
\affiliation{\institution{Duke Kunshan University}\city{Kunshan}\country{China}}
\email{dongmian.zou@duke.edu}

\author{Shixin Xu}
\authornotemark[1]
\affiliation{\institution{Duke Kunshan University}\city{Kunshan}\country{China}}
\email{shixin.xu@duke.edu}

\renewcommand{\shortauthors}{Juntang Wang et al.}

\begin{abstract}
On a given trading day, a stock-factor panel describes both how stocks differ
across factors and how each factor varies across stocks. Most stock
Transformers model relationships among stocks or across time. We introduce
\ours, which also models relationships among factors. Its factor branch
encodes all stocks' values for one factor as a token, applies self-attention
across factors, and maps the result back to one return forecast per stock.
A second branch applies attention across stocks in the same panel. A gate
combines their forecasts using trailing factor covariance and market
volatility. We evaluate both branches on NASDAQ, NYSE, and S\&P~500 with the
benchmarks' five price features and with 158 factors computed from the same
data. On the 158-factor NASDAQ panel, averaging the two forecasts yields an
information coefficient (IC) of 0.0421, compared with 0.0392 for the strongest
external baseline. The gate raises IC to 0.0443 and gives the highest
annualized excess return after costs. The equal-weight ensemble has the
higher portfolio information ratio. Both combinations also improve on the
strongest external baseline in mean IC on the other two 158-factor panels.
On the five-feature NYSE panel, the differences are within the variation
across seeds.

\end{abstract}



\begin{document}
\maketitle

\section{Introduction}
\label{sec:intro}

A stock-factor panel offers two ways to read a trading day. A row lists the
factors for one stock. A column shows where the same factor is high or low
across stocks. A Transformer that attends across rows can compare two stocks'
profiles directly. To compare the market-wide patterns of two factors, it
would have to recover those patterns from features distributed across many
stock tokens. Can a model that attends directly to the columns contribute a
different signal about future returns?

Most financial Transformers use time or stocks as the attention axis. Early
models applied sequence attention to high-frequency stock histories
\citep{dingHierarchicalMultiScaleGaussian2020}. Cross-sectional models
introduced attention among stocks and market-guided aggregation
\citep{yooAccurateMultivariateStock2021,liMASTERMarketGuidedStock2024}.
More recent work uses channel-independent stock series for pretraining or
numerical Transformers for return prediction
\citep{xiaCISTHPANPretrained2024,zhangQuantformerAttentionProfit2024,wangMachineLearningStockReturn2025}.
In these models, factors typically enter as features of a stock.

In time-series forecasting, decomposition, temporal patches, cross-dimension
attention, and variable tokens give the model direct access to different
dependencies
\citep{wuAutoformerDecomposition2021,nieTimeSeriesWorth2023,zhangCrossformerTransformerUtilizing2023,liuITransformerInvertedTransformers2024}.
For a stock-factor panel, the entries of a factor token refer to different
stocks observed on the same date. Applying attention to these tokens leaves
one state per factor, while the prediction target is one return per stock.

\ours bridges these two representations with a learned stock-aligned decoder.
One branch embeds each factor column, applies factor self-attention, and
decodes the resulting states into stock forecasts. A second branch applies
attention across stocks. Both branches use the same data and predict the same
returns. Averaging their forecasts therefore tests whether the two ways of
reading the panel make complementary errors.

The relative value of the two forecasts may also vary from day to day. We
train a gate on a later validation interval to recognize market conditions in
which to give one branch more weight. Its inputs summarize the concentration
and effective rank of the factor covariance matrix and recent market
volatility. At prediction time these quantities are available before the
target returns.

We evaluate \ours on NASDAQ, NYSE, and S\&P~500 using the benchmarks' five
price features \citep{fengTemporalRelationalRanking2019,fanStockMixerSimpleStrong2024}
and a second panel of 158 Qlib factors computed from the same price and
volume records \citep{yangQlibAIorientedQuantitative2020}. On all three
158-factor panels, the equal-weight ensemble exceeds the highest baseline IC,
and the gate improves IC further. On NASDAQ, the cross-axis pair has a mean
residual correlation of 0.541, compared with 0.823 for two independently
trained stock models. The model with the highest IC also earns the highest
annualized excess return on NASDAQ, while equal weighting delivers the
highest portfolio information ratio.

The paper makes three contributions.
\begin{itemize}
  \item \textbf{Factor tokens with a stock-level readout.} Each token
  represents one observed factor across the stock universe. A learned
  decoder maps the interacting factor tokens back to individual stocks.
  \item \textbf{A controlled test of token-axis diversity.} We compare
  averaging stock and factor forecasts with averaging two stock forecasts.
  Residual correlations and stock-aligned representation distances measure
  how the choice of axis changes the pair.
  \item \textbf{Market-dependent forecast weights.} We relate the optimal
  weight to the members' conditional error covariance and evaluate a gate
  that estimates weights from past market observations.
\end{itemize}

\section{Related Work}
\label{sec:related}

\subsection{Transformers for Forward-Return and Stock Prediction}
\label{sec:related:stock}
Financial Transformers address movement classification, stock ranking, and
return regression. For high-frequency histories, attention can account for
local price changes and gaps between trades
\citep{dingHierarchicalMultiScaleGaussian2020}. Cross-sectional models learn
asymmetric relationships among stocks \citep{yooAccurateMultivariateStock2021}
or use market information to guide aggregation within and across stocks
\citep{liMASTERMarketGuidedStock2024}. Other approaches share information
through predefined concept graphs \citep{xuHISTGraphbasedFramework2021}
or combine channel-independent pretraining with a dynamic stock hypergraph
\citep{xiaCISTHPANPretrained2024}. Quantformer constructs trading factors with
a numerical Transformer \citep{zhangQuantformerAttentionProfit2024}.
Autoformer has also been applied to multi-horizon excess returns using firm
and macroeconomic panels \citep{wangMachineLearningStockReturn2025}.
Financial language models such as FinBERT address a different input
modality, financial text \citep{araciFinBERTFinancialSentiment2019}.

\ours takes a numerical panel as input and explicitly compares the
cross-sections of observed factors. StockMixer mixes indicator, time, and
stock axes through MLP blocks \citep{fanStockMixerSimpleStrong2024}, providing
a comparison with a model that mixes axes without attention. The Temporal
Routing Adaptor learns when to use different latent trading patterns
\citep{linLearningMultipleStock2021}. In \ours the two forecasts available
to the router come from different tokenizations of the same panel.

\subsection{Time-Series Transformers}
\label{sec:related:ts}
Time-series Transformers use autocorrelation to model temporal dependencies
\citep{wuAutoformerDecomposition2021}, divide individual channels into
temporal patches \citep{nieTimeSeriesWorth2023}, or apply attention across
time and dimensions in two stages \citep{zhangCrossformerTransformerUtilizing2023}.
These choices determine which observations attention can compare directly.
In \ours, a sample is one day's panel. Each factor token spans the stocks in
that panel, and the output is a vector of their forward returns.

\subsection{Transposed and Feature-Axis Transformers}
\label{sec:related:transposed}
The closest architectural precedent is the family of inverted or feature-axis
Transformers. Axial attention factorizes attention over the axes of
multidimensional inputs \citep{hoAxialAttentionMultidimensional2019}, and
iTransformer treats each variate history as a token and applies attention
across variables \citep{liuITransformerInvertedTransformers2024}. In finance,
the Feature-Axis Transformer attends over price features for stock-movement
prediction \citep{seoExploringEfficientMarket2024}, and DPA-STIFormer models
feature changes as tokens in a double-path spatial-temporal forecaster
\citep{yanDoublePathAdaptive2024}. These models illustrate how the choice of token determines what a pairwise
attention score represents.

In the factor branch, an attention score relates two factors whose values were observed
over the same stocks on the same date. The decoder then restores the stock
dimension. This makes the output directly comparable with that of the stock
branch, allowing us to combine the forecasts and measure how their errors
co-vary.

\subsection{Factor Models, Ensembles, and Regime-Conditioned Combination}
\label{sec:related:factor}
Asset-pricing models relate the cross-section of returns to firm characteristics
\citep{famaCommonRiskFactors1993}, and machine-learning asset pricing extends
this to flexible function classes over large characteristic panels
\citep{guEmpiricalAssetPricing2020,kellyCharacteristicsAreCovariances2019},
including autoencoders and variational models that learn latent factors
\citep{guAutoencoderAssetPricing2021,duanFactorVAEProbabilisticDynamic2022} and
hypergraph-contrastive extensions that mine hidden factors beyond a prior set
\citep{duanFactorGCLHypergraphBased2025}. \ours starts from \emph{observed} factors. It asks whether the relationships
among their cross-sections add information to forecasts made from stock
tokens.

Ensemble theory relates the error of an average to the covariance of its
members' errors
\citep{kroghNeuralNetworkEnsembles1995,brownDiversityCreationMethods2005,liuEnsembleLearningNegative1999,lakshminarayananSimpleScalablePredictive2017}.
Mixtures of experts let the weights depend on the input
\citep{jacobsAdaptiveMixturesLocal1991,shazeerOutrageouslyLargeNeural2017}.
In financial forecasting, changes in market conditions can motivate such
adaptive combinations
\citep{hamiltonNewApproachEconomic1989,angRegimeChangesFinancial2012,zhaoDoubleAdaptMetalearning2023,xiangRSAPDFMRegimeShiftingAdaptive2024}.
We use the error covariance of the stock and factor forecasts to characterize
their optimal combination. Our gate estimates their weights from two
observable properties of the market, factor covariance and volatility.

\section{Method}
\label{sec:method}
\begin{figure*}[t]
\centering
\begingroup
\definecolor{ftInk}{HTML}{203244}
\definecolor{ftMute}{HTML}{4B5E70}
\definecolor{ftRow}{HTML}{245981}
\definecolor{ftCol}{HTML}{974329}
\definecolor{ftGate}{HTML}{246655}
\resizebox{\textwidth}{!}{%
\begin{tikzpicture}[
  x=1cm,y=1cm,font=\sffamily\small,text=ftInk,
  title/.style={font=\sffamily\bfseries\small,inner sep=0pt},
  body/.style={font=\sffamily\footnotesize,inner sep=0pt},
  smalltext/.style={font=\sffamily\footnotesize,text=ftMute,inner sep=0pt},
  wire/.style={line width=.8pt,line join=round,line cap=round},
  tip/.style={wire,-{Stealth[length=1.7mm,width=1.5mm]}},
  tile/.style={rounded corners=3pt,line width=.5pt},
]
\node[title,anchor=west] at (.02,4.46) {Stock--factor panel};
\node[body,anchor=west] at (.02,4.06) {$\overline{\mZ}_t\in\sR^{N\times F}$};
\begin{scope}[shift={(.42,2.04)},x=.43cm,y=.39cm]
  \fill[ftMute!5] (0,0) rectangle (5,4);
  \fill[ftRow!22] (0,2) rectangle (5,3);
  \fill[ftCol!25] (2,0) rectangle (3,4);
  \fill[ftCol!35!ftRow!35] (2,2) rectangle (3,3);
  \draw[step=1,draw=white,line width=1pt] (0,0) grid (5,4);
  \draw[ftMute!58,line width=.5pt] (0,0) rectangle (5,4);
  \draw[ftRow,line width=.75pt] (0,2) rectangle (5,3);
  \draw[ftCol,line width=.75pt] (2,0) rectangle (3,4);
  \node[smalltext,text=ftRow,anchor=east] at (-.15,2.5) {$i$};
  \node[smalltext,text=ftCol,anchor=south] at (2.5,4.16) {$j$};
\end{scope}
\node[smalltext,anchor=west] at (.02,1.72) {$N$ stocks $\times$ $F$ factors};
\filldraw[tile,fill=ftRow!5,draw=ftRow!32] (3.53,3.43) rectangle (13.69,4.97);
\filldraw[tile,fill=ftCol!6,draw=ftCol!35] (3.53,1.54) rectangle (13.69,3.08);
\draw[tip,ftRow] (2.57,3.02) -- (2.92,3.02) |- (3.53,4.2);
\draw[tip,ftCol] (1.495,2.04) -- (1.495,1.91) -- (3.16,1.91) |- (3.53,2.31);
\node[title,text=ftRow,anchor=west] at (3.79,4.59) {Encode each row};
\node[body,anchor=west] at (3.79,4.17) {One token per stock};
\foreach \i/\entry in {0/{z_{i1}},1/{z_{i2}},2/{\cdots},3/{z_{iF}}}{
  \filldraw[rounded corners=.8pt,draw=ftRow!50,fill=white,line width=.4pt]
    ({3.79+.57*\i},3.60) rectangle ++(.54,.33);
  \node[smalltext,text=ftRow] at ({4.06+.57*\i},3.765) {$\entry$};
}
\node[title,text=ftCol,anchor=west] at (3.79,2.70) {Encode each column};
\node[body,anchor=west] at (3.79,2.28) {One token per factor};
\foreach \i/\entry in {0/{z_{1j}},1/{z_{2j}},2/{\cdots},3/{z_{Nj}}}{
  \filldraw[rounded corners=.8pt,draw=ftCol!50,fill=white,line width=.4pt]
    ({3.79+.57*\i},1.71) rectangle ++(.54,.33);
  \node[smalltext,text=ftCol] at ({4.06+.57*\i},1.875) {$\entry$};
}
\foreach \y/\hue/\axis/\size in {4.20/ftRow/stocks/N,2.31/ftCol/factors/F}{
  \draw[tip,\hue] (6.26,\y) -- (6.68,\y);
  \node[title,text=\hue,anchor=west] at (6.92,{\y+.39}) {Attend over \axis};
  \begin{scope}[shift={(6.93,\y)},x=.135cm,y=.135cm]
    \fill[\hue!24] (0,-3) rectangle (4,1);
    \draw[step=1,white,line width=.5pt] (0,-3) grid (4,1);
  \end{scope}
  \node[body,anchor=west] at (7.65,{\y-.14}) {$\size\!\times\!\size$ weights};
  \draw[tip,\hue] (9.47,\y) -- (9.89,\y);
}
\node[title,text=ftRow,anchor=west] at (10.12,4.59) {Read each stock state};
\node[body,anchor=west] at (10.12,4.10) {Stock-member forecast};
\node[body,anchor=west] at (10.12,3.74) {$\widehat{\vy}^{s}_t\in\sR^N$};
\node[title,text=ftCol,anchor=west] at (10.12,2.70) {Decode to stock order};
\node[body,anchor=west] at (10.12,2.26)
  {$\mR^f_t=\mM_t(\mH_t\mW_{\mathrm{back}})^\top$};
\node[body,anchor=west] at (10.12,1.86)
  {$\widehat{\vy}^{f}_t=\mR^f_t\vw_{\mathrm{read}}$};
\filldraw[tile,fill=ftGate!5,draw=ftGate!36] (.02,-.08) rectangle (13.69,1.16);
\node[title,text=ftGate,anchor=west] at (.28,.78) {Trailing market state};
\node[smalltext,anchor=west] at (.28,.37) {Available through date $t$};
\draw[draw=ftGate!35] (3.53,.14) -- (3.53,.94);
\node[body,anchor=west] at (3.79,.77) {60-day factor covariance};
\node[smalltext,anchor=west] at (3.79,.35) {Concentration and effective rank};
\node[body,anchor=west] at (7.67,.77) {20-day index returns};
\node[smalltext,anchor=west] at (7.67,.35) {Market volatility};
\draw[tip,ftGate] (10.44,.54) -- (10.87,.54);
\node[title,text=ftGate,anchor=west] at (11.09,.77) {Regime gate};
\node[smalltext,anchor=west] at (11.09,.35) {Softmax weights};
\filldraw[tile,fill=ftInk!4,draw=ftInk!35] (14.41,1.54) rectangle (17.60,4.97);
\draw[tip,ftRow] (13.69,4.20) -- (14.41,4.20);
\draw[tip,ftCol] (13.69,2.31) -- (14.41,2.31);
\draw[tip,ftGate] (13.69,.54) -| (16.01,1.54);
\node[smalltext,text=ftGate,anchor=south] at (14.84,.63) {$(g_t^s,g_t^f)$};
\node[title] at (16.01,4.56) {Combine forecasts};
\node[body] at (16.01,3.88) {$\widehat{\vy}_t =
    \textcolor{ftRow}{g_t^s\widehat{\vy}^{s}_t} +
    \textcolor{ftCol}{g_t^f\widehat{\vy}^{f}_t}$};
\draw[ftInk!25] (14.75,3.42) -- (17.26,3.42);
\node[body,align=center] at (16.01,2.78) {One forward-return\\forecast per stock};
\node[smalltext] at (16.01,1.94) {$g_t^s+g_t^f=1$};
\end{tikzpicture}%
}
\endgroup
\caption{FToken reads a daily panel along two axes. The highlighted row gives
stock $i$'s factor profile. The highlighted column gives factor $j$'s values
across stocks. After attention, the factor branch decodes its tokens back into stock order.
Trailing covariance and volatility determine the weights on the two forecasts.}
\Description{Parallel paths read a stock-by-factor matrix by row and by column.
Stock tokens attend to stocks and yield stock-level forecasts. Factor tokens
attend to factors, then a dense decoder and a readout yield a forecast for each
stock. Independently, factor covariance concentration, effective rank and
index volatility feed a softmax gate. The final forecast is the weighted sum
of the two forecasts.}
\label{fig:architecture}
\end{figure*}
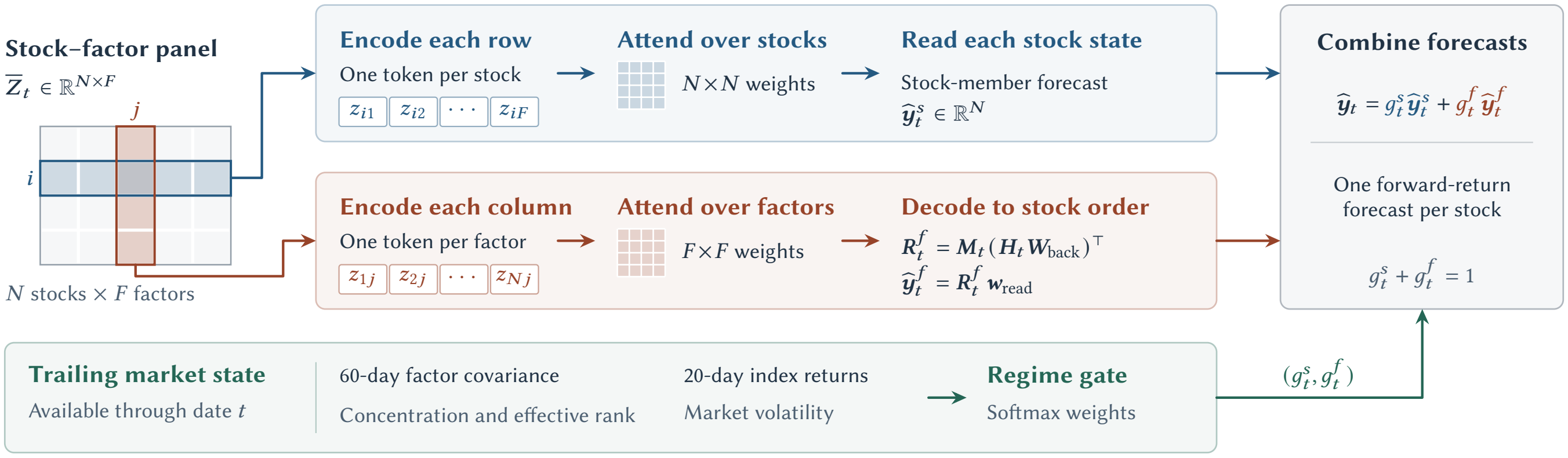

\subsection{Panel Setup and an Axis-Matched Stock Member}
\label{sec:setup}
On date $t$, the standardized factor panel is
$\mZ_t\in\sR^{N\times F}$, where columns are observed factors and rows follow a
canonical benchmark stock index. Let $\vm_t\in\{0,1\}^{N}$ be the
point-in-time availability mask, $n_t=\sum_i m_{t,i}$, and
$\mM_t=\operatorname{diag}(\vm_t)$. Define the active-set centering operator
$\mP_t=\mM_t-\vm_t\vm_t^{\top}/n_t$. The target $\vy_t\in\sR^N$ contains
cross-sectionally normalized next-period returns in the same canonical order.
Losses and metrics are evaluated only on active entries.

The factor encoder and decoder assign weights to fixed stock identities.
We therefore keep each benchmark's pre-filtered pool in a fixed ticker order
($N=1026$ for NASDAQ, $1737$ for NYSE, and $474$ for S\&P~500).
We fill missing entries with the daily cross-sectional median and then apply
$\vm_t$ to mask non-trading stock-days. We denote the resulting panel by
$\overline{\mZ}_t=\mathcal M(\mZ_t,\vm_t)$.

A conventional stock-axis model treats row $i$ as one token and produces
\begin{equation}
  \widehat{\vy}^{s}_t=f^{s}_{\vtheta_s}(\overline{\mZ}_t,\vm_t).
  \label{eq:stockaxis}
\end{equation}
The stock Transformer in \cref{eq:stockaxis} shares its depth, hidden width,
head count, optimizer, loss, and early-stopping rule with the factor member
(\cref{tab:protocol}). The projections specific to each axis give the
branches different parameter counts (\cref{tab:ablation}).

\subsection{Factors as Tokens and Re-Transposed Readout}
\label{sec:factor-member}
Stock attention compares rows of the panel. The factor branch instead
projects each column into a token of dimension $d$.
\begin{equation}
  \mH_t^{(0)}=\overline{\mZ}_t^{\top}\mW_{\mathrm{in}}
  \in\sR^{F\times d},
  \qquad \mW_{\mathrm{in}}\in\sR^{N\times d}.
  \label{eq:factortokens}
\end{equation}
For head $a$ of width $d_a$ with $\sum_a d_a=d$, learned projections form
$\mQ_a$, $\mK_a$, and $\mV_a$. Multi-head self-attention
\citep{vaswaniAttentionAllYou2017} mixes the $F$ factor tokens as follows.
\begin{equation}
  \mH_t^{(\ell)}=\mathop{\mathrm{Concat}}_a\!\left[
  \softmax\!\left(\frac{\mQ_a\mK_a^{\top}}{\sqrt{d_a}}\right)\mV_a
  \right]\mW_O\in\sR^{F\times d},
  \label{eq:factorattn}
\end{equation}
where $\mW_O\in\sR^{d\times d}$ is the output projection. Each block combines
this attention sublayer with residual connections, layer normalization, and
a feed-forward sublayer. Let $\mH_t$ denote the output of the final block.
The decoder restores one row per stock, then combines the factors within each
row to obtain a forecast.
\begin{equation}
  \mR_t^f=\mM_t\bigl(\mH_t\mW_{\mathrm{back}}\bigr)^{\top}
  \in\sR^{N\times F},\qquad
  \widehat{\vy}^{f}_t=\mR_t^f\vw_{\mathrm{read}},
  \label{eq:factorreadout}
\end{equation}
where $\mW_{\mathrm{back}}\in\sR^{d\times N}$ and
$\vw_{\mathrm{read}}\in\sR^F$. The factor member's parameters
$\vtheta_f$ comprise $\mW_{\mathrm{in}}$, the attention blocks,
$\mW_{\mathrm{back}}$, and $\vw_{\mathrm{read}}$.
\Cref{fig:architecture} shows how the two branches predict the same stock
returns from different views of the panel. In the results tables, \ours
denotes the complete system and \emph{\ours member} denotes the factor
branch alone.

\paragraph{Stock identity and parameter count.}
Each row of $\mW_{\mathrm{in}}$ and each column of $\mW_{\mathrm{back}}$
correspond to a particular stock. Consequently the dense implementation
requires a fixed universe and a consistent stock order. We test the role of
these parameters with a stock-axis control of comparable size and a
stock-query cross-attention readout that replaces the dense $d\times N$
decoder. We also examine sensitivity to masks and constituent turnover.

\subsection{Conditional Complementarity and the Optimal Weight}
\label{sec:complementarity}
For one date, let the active-set centered member errors be
$\ve_t^s=\mP_t(\vy_t-\widehat{\vy}_t^s)$ and
$\ve_t^f=\mP_t(\vy_t-\widehat{\vy}_t^f)$. Consider a convex ensemble
\begin{equation}
  \widehat{\vy}_t(g)=g\widehat{\vy}_t^s+(1-g)\widehat{\vy}_t^f,
  \qquad 0\le g\le 1.
  \label{eq:convexensemble}
\end{equation}
Conditioned on observable regime information $\vr_t$, define per-stock moments
\begin{equation}
\begin{aligned}
  v_s(\vr_t)&=\E[\|\ve_t^s\|_2^2/n_t\mid\vr_t],\\
  v_f(\vr_t)&=\E[\|\ve_t^f\|_2^2/n_t\mid\vr_t],\\
  c(\vr_t)&=\E[(\ve_t^s)^{\top}\ve_t^f/n_t\mid\vr_t],
\end{aligned}
\label{eq:conditionalmoments}
\end{equation}
and let
$\mathcal R(g\mid\vr_t)
=\E\bigl[\|\mP_t(\vy_t-\widehat{\vy}_t(g))\|_2^2/n_t\mid\vr_t\bigr]$
denote the conditional ensemble risk. The following specializes the classical
error-covariance analysis of ensembles
\citep{kroghNeuralNetworkEnsembles1995,brownDiversityCreationMethods2005} to
the regime-conditional two-member case.

\begin{proposition}[Conditional optimal routing]
\label{prop:optimalgate}
Let $D(\vr_t)=v_s(\vr_t)+v_f(\vr_t)-2c(\vr_t)>0$. The unconstrained
squared-error minimizer is
\begin{equation}
  g^*(\vr_t)=\frac{v_f(\vr_t)-c(\vr_t)}{D(\vr_t)},
  \label{eq:optimalweight}
\end{equation}
and the convex optimum is $\clip(g^*(\vr_t),0,1)$. For any unconstrained
$g$, the conditional excess risk satisfies
\begin{equation}
  \mathcal R(g\mid\vr_t)-\mathcal R(g^*\mid\vr_t)
  =D(\vr_t)\bigl(g-g^*(\vr_t)\bigr)^2.
  \label{eq:excessrisk}
\end{equation}
\end{proposition}

The derivation is in \cref{app:proof}. When the two members make different
errors, a sufficiently low covariance $c$ can give the ensemble lower risk
than either member alone. A market-dependent weight can improve on the best
fixed weight when the observable market state helps predict changes in the
optimal combination.

To measure how closely the members' errors align, we compute the daily
cross-sectional residual correlation
\begin{equation}
  \rho_t=\frac{(\ve_t^s)^{\top}\ve_t^f}
  {\|\ve_t^s\|_2\|\ve_t^f\|_2},
  \label{eq:residualcorr}
\end{equation}
after standardizing each forecast to unit cross-sectional scale and centering
the errors on the active stocks. This gives one correlation per date. We
aggregate the daily values using the Fisher $z$ transform.

\subsection{Stock-Aligned Representation Geometry}
\label{sec:geometry}
To compare the branches before they produce their forecasts, we measure the
overlap between their learned representations in stock space. Let
$\mR_t^s\in\sR^{N\times d_s}$ be the stock member's penultimate stock-aligned
representation after masking, where $d_s$ is that member's hidden width, and
let $\mR_t^f\in\sR^{N\times F}$ be the factor
member's decoded representation from \cref{eq:factorreadout}. For a rolling
window $\mathcal W_t$, concatenate centered representations along their feature
axes,
\begin{equation}
  \mA_t^s=[\mP_\tau\mR_\tau^s]_{\tau\in\mathcal W_t},\qquad
  \mA_t^f=[\mP_\tau\mR_\tau^f]_{\tau\in\mathcal W_t},
\end{equation}
and let $\mU_t^s,\mU_t^f\in\sR^{N\times k}$ contain their leading $k$ left singular
vectors. With principal angles $\theta_{t,j}$ between the two column spaces
\citep{bjorckNumericalMethodsComputing1973,golubMatrixComputations2013}, define
\begin{equation}
  \kappa_t=1-\frac{1}{k}\sum_{j=1}^{k}\cos^2\theta_{t,j},
  \qquad
  \cos\theta_{t,j}=s_j\!\left((\mU_t^s)^{\top}\mU_t^f\right),
  \label{eq:kappa}
\end{equation}
where $s_j(\cdot)$ is the $j$th singular value. This squared normalized chordal
distance is near zero for overlapping subspaces and larger for distinct ones. We compare $\kappa_t$ with the residual correlation and the IC gain from
averaging the forecasts. We also compute linear centered kernel alignment
(CKA) \citep{kornblithSimilarityNeuralNetwork2019} as a second measure of
representation similarity, since the principal angles depend on the layer,
window, and rank $k$.

\subsection{Past-Only Regime Features and Gate}
\label{sec:gate}
Market regimes can differ in both volatility and the structure of factor
covariance \citep{hamiltonNewApproachEconomic1989,angRegimeChangesFinancial2012}.
To capture these differences at the prediction date, the gate uses covariance
concentration, normalized effective rank, and trailing market volatility.
Over a trailing window of length $L$, we compute
\begin{equation}
  \mC_t=\frac{1}{L}\sum_{\tau=t-L+1}^{t}
  \frac{\overline{\mZ}_{\tau}^{\top}\mP_\tau
        \overline{\mZ}_{\tau}}{\max(n_\tau-1,1)}
\end{equation}
as the mean cross-sectional factor covariance available at the prediction timestamp.
If $\lambda_{t,j}$ are its eigenvalues in descending order and
$p_{t,j}=\lambda_{t,j}/\sum_{\ell}\lambda_{t,\ell}$, the regime vector is
\begin{equation}
  \vr_t=\left[
  p_{t,1},\;
  \frac{\exp(-\sum_j p_{t,j}\log(p_{t,j}+\epsilon))}{F},\;
  \nu^{\mathrm{mkt}}_t
  \right],
\end{equation}
where $\epsilon>0$ is a small numerical-stability constant and
$\nu^{\mathrm{mkt}}_t$ is the trailing 20-day standard deviation of the
benchmark-index return. We average factor covariance over $L=60$ trading days. The largest eigenvalue
share measures the fraction of variation along the leading factor direction.
The normalized effective rank measures how widely variation is spread across
directions.

A small network maps $\vr_t$ to two weights that sum to one, as in a
two-expert mixture \citep{jacobsAdaptiveMixturesLocal1991,shazeerOutrageouslyLargeNeural2017}.
The resulting forecast is
\begin{equation}
  (g_t^s,g_t^f)=\softmax(h_{\vtheta_g}(\vr_t)),\qquad
  \widehat{\vy}_t=g_t^s\widehat{\vy}_t^s+g_t^f\widehat{\vy}_t^f.
  \label{eq:gate}
\end{equation}
We compare the learned weights with equal weights, a static weight selected
on validation data, and a gate based on trailing forecast performance. An
oracle computed from realized member errors provides a hindsight reference
for \cref{eq:optimalweight}.

\subsection{Chronological Training}
\label{sec:training}
Training proceeds in two stages (\cref{alg:training}). First, both branches
minimize mean squared error (MSE) on cross-sectionally normalized returns.
This normalization puts the daily targets on a common scale for training and
for evaluating Pearson IC. We then freeze the branches and use their forecasts
on a later validation interval to train the router. This holds the two
forecast series fixed while the router learns their weights. We compare this
procedure with joint end-to-end training in \cref{sec:exp:ablation}. All
preselected seeds enter the evaluation.

\subsection{Complexity}
\label{sec:complexity}
Changing the token axis changes where computation is spent. For hidden width
$d$, a stock-attention layer requires $O(N^2d+Nd^2)$ work, while the factor
branch requires $O(F^2d+NFd+Fd^2)$ for transposed encoding, factor attention,
and decoding. With $F=158$ and roughly $500$ to $1{,}700$ stocks, the factor branch forms
a smaller attention matrix. Its dense encoder and decoder each require $Nd$
parameters. \Cref{sec:exp:ablation} reports the measured parameter counts,
training time, peak memory, and gate latency.

\section{Experiments}
\label{sec:experiments}

\subsection{Research Questions}
\label{sec:exp:rq}
We ask how much the choice of token axis contributes to prediction and how
much a market-dependent weight adds once both forecasts are available.
\begin{enumerate}
  \item How do the branches perform individually and in combination, as
  measured by IC and portfolio returns after costs?
  \item How do their errors and stock-aligned representations compare with
  those of two independently trained stock models?
  \item Can past factor covariance and volatility help the gate improve on
  equal weights and on a static weight tuned on validation data?
  \item How sensitive are the results to the feature set, seed, subperiod,
  readout, gate window, and trading costs, and how much computation does each
  branch require?
\end{enumerate}

\subsection{Data, Labels, and Leakage Controls}
\label{sec:data}
All evaluations use the public NASDAQ, NYSE, and S\&P~500 stock-prediction
benchmarks: the NASDAQ and NYSE universes of
\citet{fengTemporalRelationalRanking2019} (2013 to 2017, stocks filtered for
near-complete trading histories) and the S\&P~500 set distributed with
StockMixer \citep{fanStockMixerSimpleStrong2024} (474 stocks, 2016 to 2022).
The benchmarks split the dates into training, route-validation, and test
intervals, in that order. Their lengths are 756/252/237 trading days for
NASDAQ and NYSE and 1006/253/352 for S\&P~500. We take the raw data from the
source repositories and evaluate two feature sets on the same universes and
dates.

\paragraph{Native five-feature tier.}
The native protocol uses five normalized price features per stock-day and a
one-day raw return label. The released evaluator reports IC, the information
ratio of the daily IC series (ICIR), precision at ten, and the annualized
Sharpe ratio of a daily top-five portfolio before costs. We label the second
metric ICIR throughout, following the evaluator's calculation. The published
tables call it rank IC.

\paragraph{158-factor tier.}
We compute the 158 Qlib Alpha158 factors
\citep{yangQlibAIorientedQuantitative2020} from the same public OHLCV records.
The target is the one-day forward close-to-close return with a one-day lag,
normalized across stocks on each date. We run every model on the benchmark
splits under a common protocol. We evaluate IC, ICIR, RankIC, and RankICIR
and compare top-$K$ dropout portfolios after costs with each universe's index.
The two tiers use different labels, portfolio rules, and evaluators, so we
make method comparisons within each tier.

\paragraph{Information available at prediction time.}
Factor construction, cross-sectional normalization, missing-value handling,
and all covariance and volatility windows use information available by the
prediction timestamp. We fit preprocessing and select architectures, gate
features, and portfolio settings before test evaluation. A saved configuration
specifies these settings. We retain the benchmarks' fixed stock pools for
comparability and stratify results by the proportion of stocks masked on each
date.

\subsection{Baselines and Controlled Variants}
\label{sec:baselines}
On the native five-feature tier the comparison set follows the established
protocol: LSTM, ALSTM \citep{qinDualstageAttentionbasedRecurrent2017}, GAT
\citep{velickovicGraphAttentionNetworks2018}, STHAN-SR
\citep{sawhneyStockSelectionSpatiotemporal2021}, ESTIMATE
\citep{huynhEfficientIntegrationMultiOrder2023}, and StockMixer
\citep{fanStockMixerSimpleStrong2024}, all re-run on the shared public
datasets. On the 158-factor tier the baselines cover linear, tree, temporal,
relational, and ensemble families that accept a generic feature panel:
Linear, MLP, LightGBM \citep{keLightGBMHighlyEfficient2017}, ALSTM,
Transformer \citep{vaswaniAttentionAllYou2017}, GAT, DoubleEnsemble
\citep{zhangDoubleEnsembleNew2020}, and StockMixer extended to the
158-feature input. We use official or protocol-compatible implementations.
Tuning uses pre-test data, and temporal models receive histories through the
same prediction timestamp. HIST requires concept graphs
\citep{xuHISTGraphbasedFramework2021}, and MASTER uses market-guided feature
selection specific to its index context \citep{liMASTERMarketGuidedStock2024}.
The public panels lack these auxiliary inputs.

To isolate the choice of axis, we compare the stock branch with the factor
branch, train a second stock model with an independent seed, and average the
two stock forecasts. We also replace factor self-attention with an MLP and
test a stock model with comparable identity-specific capacity and parameter
count. For the paired forecasts we compare equal weights, a validation-tuned
static weight, a trailing-performance gate, and the full regime gate. The
hindsight oracle weights the members using their realized errors.
\Cref{tab:ablation} reports parameter counts and compares the dense decoder
in \cref{sec:factor-member} with a stock-query cross-attention readout.

\subsection{Metrics, Backtest, and Uncertainty}
\label{sec:exp:metrics}
For each test date we compute the Pearson correlation between forecasts and
returns across active stocks. IC reports the mean of these daily correlations. RankIC uses the corresponding rank correlations.
ICIR and RankICIR divide the time mean of each daily series by its standard
deviation. These four metrics apply to the 158-factor tier. The five-feature
tier follows its native evaluator (\cref{sec:data}). To compare branches we
measure daily residual correlation $\rho_t$ (\cref{eq:residualcorr}),
principal-angle distance $\kappa_t$ (\cref{eq:kappa}), CKA, and the IC gain
from averaging forecasts. For the router we also measure weight entropy and
the IC difference from the hindsight oracle.

For every model in the 158-factor tier, we hold $K=50$ stocks and allow $5$
replacements per day under a top-$K$ dropout rule. Trades execute at the close
with a one-way cost of 5 basis points on purchases and sales. The benchmarks
are the NASDAQ Composite, NYSE Composite, and S\&P~500, respectively. We
exclude non-trading stock-days, annualize using 252 trading days, and report
net annualized excess return, portfolio information ratio (IR), maximum
drawdown, and two-sided daily turnover. \Cref{sec:exp:ablation} varies the
trading cost and $K$.

Each stochastic model uses ten independent seeds (0--9). Linear and LightGBM
are deterministic and run once. The tables report seed means and, where shown,
standard deviations. We compute paired 95\% confidence intervals for
differences in daily IC and portfolio returns using a moving-block bootstrap
with 20-day blocks. The intervals apply to individual comparisons without
adjustment for multiplicity.

\subsection{Implementation Details}
\label{sec:exp:implementation}
Both branches use the same hidden width, depth, head count, dropout search
space, optimizer, batch construction, loss, patience, and maximum number of
epochs. We use the route-validation interval for early stopping and gate
fitting and select hyperparameters on its ICIR. Pre-registered grids cover the
covariance window $L$, principal-angle rank $k$, and gate width. We fix the
backtest configuration before evaluating the test period. Runtime comparisons
use the same GPU, software, precision, and number of data-loader workers
(\cref{tab:protocol}). \Cref{app:audit} describes how we retain stock
identities across dates.

\subsection{Main Results}
\label{sec:exp:main}
\begin{table*}[t]
\centering
\caption{Predictive performance on the 158-factor NASDAQ and S\&P~500 panels
(\cref{sec:data}). We evaluate all methods under the same protocol. Entries
give means and standard deviations over ten seeds for stochastic methods.
Linear and LightGBM are deterministic. Shading marks the three highest values
in each column.}
\label{tab:predictive}
\small
\setlength{\tabcolsep}{6.0pt}
\renewcommand{\arraystretch}{0.94}
\resizebox{\textwidth}{!}{%
\begin{tabular}{lcccccccc}
\toprule
& \multicolumn{4}{c}{NASDAQ} & \multicolumn{4}{c}{S\&P~500}\\
\cmidrule(lr){2-5}\cmidrule(lr){6-9}
Method & IC & ICIR & RankIC & RankICIR & IC & ICIR & RankIC & RankICIR\\
\midrule
Linear & 0.0298 & 0.2173 & 0.0341 & 0.2363 & 0.0243 & 0.1756 & 0.0281 & 0.1958 \\
MLP & 0.0301\s{0.0019} & 0.2215\s{0.0136} & 0.0352\s{0.0020} & 0.2759\s{0.0141} & 0.0259\s{0.0019} & 0.1911\s{0.0136} & 0.0304\s{0.0020} & 0.2343\s{0.0139} \\
LightGBM & 0.0371 & 0.3010 & 0.0359 & 0.2929 & 0.0327 & 0.2586 & 0.0318 & 0.2629 \\
ALSTM & 0.0322\s{0.0030} & 0.2550\s{0.0223} & 0.0401\s{0.0024} & 0.3236\s{0.0192} & 0.0281\s{0.0020} & 0.2079\s{0.0139} & 0.0349\s{0.0022} & 0.2741\s{0.0171} \\
Transformer & 0.0264\s{0.0025} & 0.1965\s{0.0188} & 0.0355\s{0.0027} & 0.2716\s{0.0194} & 0.0262\s{0.0044} & 0.1901\s{0.0322} & 0.0322\s{0.0041} & 0.2478\s{0.0297} \\
GAT & 0.0342\s{0.0014} & 0.2743\s{0.0111} & 0.0422\s{0.0015} & 0.3373\s{0.0109} & 0.0304\s{0.0013} & 0.2376\s{0.0101} & 0.0377\s{0.0012} & 0.3077\s{0.0089} \\
DoubleEnsemble & 0.0392\s{0.0028} & 0.3296\s{0.0247} & 0.0381\s{0.0028} & 0.3251\s{0.0227} & 0.0356\s{0.0014} & 0.3015\s{0.0124} & 0.0344\s{0.0014} & 0.2888\s{0.0112} \\
StockMixer & 0.0345\s{0.0033} & 0.2771\s{0.0254} & 0.0403\s{0.0036} & 0.3261\s{0.0291} & 0.0311\s{0.0032} & 0.2491\s{0.0246} & 0.0362\s{0.0037} & 0.2948\s{0.0301} \\
\midrule
Stock-axis member & 0.0334\s{0.0020} & 0.2718\s{0.0166} & 0.0398\s{0.0015} & 0.3209\s{0.0111} & 0.0302\s{0.0028} & 0.2438\s{0.0231} & 0.0361\s{0.0028} & 0.2907\s{0.0207} \\
Same-axis stock ensemble & 0.0352\s{0.0015} & 0.2871\s{0.0120} & 0.0421\s{0.0017} & 0.3457\s{0.0135} & 0.0318\s{0.0021} & 0.2645\s{0.0171} & 0.0382\s{0.0021} & 0.3165\s{0.0169} \\
Stock-axis capacity control & 0.0338\s{0.0020} & 0.2582\s{0.0159} & 0.0405\s{0.0023} & 0.3293\s{0.0168} & 0.0306\s{0.0016} & 0.2344\s{0.0127} & 0.0368\s{0.0019} & 0.2864\s{0.0133} \\
Factor-token MLP & 0.0287\s{0.0029} & 0.2180\s{0.0209} & 0.0331\s{0.0034} & 0.2546\s{0.0256} & 0.0244\s{0.0034} & 0.1834\s{0.0243} & 0.0283\s{0.0038} & 0.2188\s{0.0288} \\
FToken member & 0.0311\s{0.0018} & 0.2360\s{0.0137} & 0.0369\s{0.0016} & 0.2987\s{0.0124} & 0.0271\s{0.0018} & 0.2057\s{0.0137} & 0.0323\s{0.0014} & 0.2521\s{0.0105} \\
Equal-weight dual-axis & \third{0.0421\s{0.0014}} & \second{0.3608\s{0.0116}} & \third{0.0492\s{0.0014}} & \third{0.4186\s{0.0120}} & \second{0.0389\s{0.0017}} & \first{0.3243\s{0.0137}} & \second{0.0451\s{0.0015}} & \second{0.3886\s{0.0131}} \\
Static validation-optimal & \second{0.0426\s{0.0017}} & \third{0.3490\s{0.0143}} & \second{0.0497\s{0.0019}} & \second{0.4322\s{0.0154}} & \third{0.0381\s{0.0018}} & \third{0.3083\s{0.0150}} & \third{0.0439\s{0.0017}} & \third{0.3768\s{0.0136}} \\
\textbf{FToken full (regime gate)} & \first{0.0443\s{0.0018}} & \first{0.3651\s{0.0147}} & \first{0.0517\s{0.0011}} & \first{0.4414\s{0.0089}} & \first{0.0405\s{0.0019}} & \second{0.3230\s{0.0150}} & \first{0.0468\s{0.0016}} & \first{0.3968\s{0.0129}} \\
\bottomrule
\end{tabular}}
\end{table*}

\begin{table*}[t]
\centering
\caption{Portfolio performance after costs on the 158-factor NASDAQ and
S\&P~500 panels under the top-$K$ dropout rule (\cref{sec:exp:metrics}).
Ann.\ Ret.\ denotes annualized excess return. Shading marks the three highest
returns and information ratios and the three drawdowns closest to zero.
Entries give means and standard deviations over ten seeds for stochastic
methods. LightGBM is deterministic. The five external baselines have the
highest NASDAQ ICs in \cref{tab:predictive} among the external methods.}
\label{tab:portfolio}
\small
\setlength{\tabcolsep}{6.5pt}
\renewcommand{\arraystretch}{0.94}
\resizebox{\textwidth}{!}{%
\begin{tabular}{lcccccccc}
\toprule
& \multicolumn{4}{c}{NASDAQ} & \multicolumn{4}{c}{S\&P~500}\\
\cmidrule(lr){2-5}\cmidrule(lr){6-9}
Method & Ann. Ret. & IR & MDD & Turnover & Ann. Ret. & IR & MDD & Turnover\\
\midrule
LightGBM & 0.0685 & 1.09 & -0.0747 & 0.188 & 0.0512 & 0.74 & -0.0789 & 0.184 \\
ALSTM & 0.0398\s{0.0232} & 0.68\s{0.43} & -0.0805\s{0.0161} & 0.174\s{0.005} & 0.0261\s{0.0198} & 0.45\s{0.37} & -0.0725\s{0.0145} & 0.188\s{0.006} \\
GAT & 0.0596\s{0.0118} & 0.82\s{0.18} & -0.0824\s{0.0165} & 0.175\s{0.010} & 0.0468\s{0.0115} & 0.62\s{0.16} & -0.0852\s{0.0170} & 0.182\s{0.011} \\
DoubleEnsemble & 0.0608\s{0.0204} & 1.00\s{0.36} & -0.0869\s{0.0174} & 0.186\s{0.005} & 0.0524\s{0.0132} & \third{1.04\s{0.28}} & \first{-0.0557\s{0.0111}} & 0.176\s{0.007} \\
StockMixer & 0.0471\s{0.0225} & 0.92\s{0.47} & -0.0751\s{0.0150} & 0.173\s{0.006} & 0.0348\s{0.0258} & 0.53\s{0.42} & -0.0890\s{0.0178} & 0.187\s{0.011} \\
\midrule
Stock-axis member & 0.0545\s{0.0180} & 0.81\s{0.29} & -0.0706\s{0.0141} & 0.172\s{0.011} & 0.0402\s{0.0291} & 0.55\s{0.43} & -0.0805\s{0.0161} & 0.176\s{0.008} \\
FToken member & 0.0415\s{0.0155} & 0.77\s{0.31} & \first{-0.0563\s{0.0113}} & 0.191\s{0.005} & 0.0256\s{0.0173} & 0.36\s{0.26} & -0.1037\s{0.0207} & 0.184\s{0.006} \\
Equal-weight dual-axis & \second{0.0754\s{0.0113}} & \first{1.34\s{0.22}} & \second{-0.0575\s{0.0115}} & 0.178\s{0.005} & \first{0.0702\s{0.0145}} & \first{1.35\s{0.30}} & \third{-0.0720\s{0.0144}} & 0.183\s{0.006} \\
Static validation-optimal & \third{0.0728\s{0.0150}} & \second{1.29\s{0.29}} & \third{-0.0620\s{0.0124}} & 0.181\s{0.010} & \third{0.0678\s{0.0190}} & \second{1.21\s{0.37}} & \second{-0.0583\s{0.0117}} & 0.188\s{0.007} \\
\textbf{FToken full (regime gate)} & \first{0.0842\s{0.0128}} & \third{1.12\s{0.18}} & -0.0959\s{0.0192} & 0.198\s{0.002} & \second{0.0701\s{0.0174}} & 0.99\s{0.27} & -0.0775\s{0.0155} & 0.196\s{0.003} \\
\bottomrule
\end{tabular}}
\end{table*}

\paragraph{Combining the two axes.}
On the 158-factor NASDAQ panel, mean IC is 0.0334\s{0.0020} for the
stock branch and 0.0311\s{0.0018} for the factor branch. Averaging their predictions raises
IC to 0.0421\s{0.0014}, a gain of 0.0087 over the stock branch (paired 95\%
CI [0.0061, 0.0113]). Averaging two independently trained stock branches
reaches 0.0352. The strongest external baseline, DoubleEnsemble, reaches
0.0392, giving the cross-axis average an advantage of 0.0029 (95\% CI
[0.0006, 0.0052]). On S\&P~500 the individual branches reach 0.0302 and
0.0271. Their average reaches 0.0389 versus 0.0356 for DoubleEnsemble, a
difference of 0.0033 (95\% CI [0.0009, 0.0057], \cref{tab:predictive}).

\paragraph{Adapting the weights.}
The regime gate adds 0.0022 IC over equal weighting on NASDAQ (95\% CI
[0.0007, 0.0037]) and 0.0016 on S\&P~500 (95\% CI [0.0002, 0.0030]).
A static weight tuned on validation data reaches 0.0381 on S\&P~500, compared
with 0.0389 for equal weights.

\paragraph{Portfolio returns and turnover.}
The gate gives the highest annualized excess return after costs on NASDAQ
(0.0842). Its portfolio turns over 0.198 of its value per day, close to the
strategy's cap, compared with 0.178 for equal weights. The gated portfolio
also has higher tracking error. Equal weighting yields the highest portfolio
IR (1.34). On S\&P~500 the gated and equal-weight portfolios earn nearly
identical annualized excess returns (0.0701 and 0.0702), with IRs of 0.99
and 1.35, respectively. The factor branch has the smallest maximum drawdown
on NASDAQ and DoubleEnsemble the smallest on S\&P~500 (\cref{tab:portfolio}).

\subsection{Cross-Universe and Native-Protocol Results}
\label{sec:exp:cross}
\begin{table}[t]
\centering
\caption{Results on NASDAQ under the native five-feature protocol
(\cref{sec:data}). P@10 denotes precision at ten and SR the Sharpe ratio.
Shading marks the three highest seed means in each column. \Cref{tab:usfull}
gives results for all three universes.}
\label{tab:us}
\small
\setlength{\tabcolsep}{4.2pt}
\begin{tabular}{lcccc}
\toprule
Method & IC & ICIR & P@10 & SR\\
\midrule
LSTM & 0.031 & 0.348 & 0.516 & 0.87 \\
ALSTM & 0.034 & 0.365 & 0.520 & 0.96 \\
GAT & 0.036 & 0.381 & 0.527 & 1.19 \\
STHAN-SR & 0.040 & 0.443 & 0.541 & 1.38 \\
ESTIMATE & 0.038 & 0.439 & 0.542 & 1.33 \\
StockMixer & \third{0.042} & \third{0.487} & \third{0.543} & \third{1.42} \\
\midrule
Stock-axis member & 0.039 & 0.428 & 0.536 & 1.31 \\
FToken member & 0.033 & 0.361 & 0.518 & 1.02 \\
Equal-weight dual-axis & \second{0.044} & \second{0.492} & \second{0.546} & \second{1.51} \\
\textbf{FToken full (regime gate)} & \first{0.046} & \first{0.507} & \first{0.549} & \first{1.58} \\
\bottomrule
\end{tabular}
\end{table}

NYSE has the largest stock universe. On its 158-factor panel, an equal-weight
combination reaches IC 0.0301\s{0.0021}, compared with 0.0281\s{0.0017} for
DoubleEnsemble, the strongest external baseline. The gate reaches
0.0316\s{0.0022}. Equal weights give a slightly higher annualized excess return
than the gate (\cref{tab:nyse}).

With the five native price features, both dual-axis combinations exceed every
individual model in mean IC on all three universes (\cref{tab:us,tab:usfull}).
On NASDAQ, the stock and factor branches reach 0.039 and 0.033, respectively,
and the gate reaches 0.046. On NYSE, the gated model reaches 0.032 against
0.030 for ESTIMATE, a difference within the variation across seeds. The
highest NYSE Sharpe ratio belongs to StockMixer, and the highest S\&P~500
precision at ten to ESTIMATE. Across both feature sets, the factor branch
contributes most clearly through its effect on the combined forecast.

\subsection{Why Does the Ensemble Improve?}
\label{sec:exp:mechanism}
On the NASDAQ 158-factor panel, the mean residual correlation is 0.541 for the
stock--factor pair and 0.823 for two independently seeded stock models
(\cref{tab:routing}). The cross-axis pair also spans more distinct
representation subspaces. Its principal-angle distance is 0.476 versus 0.208
for the stock pair, and its linear CKA is 0.41 versus 0.78.

We next ask whether periods with less aligned errors also yield larger gains
from averaging. We use 60-day windows at a stride of 20 days, giving nine
windows per seed. Across ten seeds (90 observations), a one-standard-deviation
decrease in Fisher-transformed residual correlation is associated with an
additional 0.14 percentage points of equal-weight IC gain (95\% CI
[0.08, 0.20]). The regression controls for the difference in member quality
and clusters standard errors by calendar window. Subspace distance increases
with decreasing residual correlation (Spearman $-0.62$). Once residual correlation
is included, the association between distance and gain is not statistically significant. Comparing gains predicted from the empirical error
moments in \cref{prop:optimalgate} with observed window-level gains gives a
slope of 0.91 and $R^2$ of 0.69.

\subsection{Does the Gate Track Market State?}
\label{sec:exp:gate}
We group NASDAQ test dates by covariance concentration, effective rank, and
volatility and examine the factor weight and the corresponding IC gain. In
the lowest and highest covariance-concentration quintiles, the mean factor
weights are 0.33 and 0.58, respectively. The IC gains over equal weights are
0.0009 and 0.0044. The middle quintiles show no monotonic trend. Mean
gate entropy is 0.64 out of a maximum of 0.69.

\Cref{tab:routing} compares routers on the same forecast pairs. The full gate
improves IC over the stock branch by 0.0109, exceeding the gains for equal
weights, the tuned static weight, and the trailing-performance gate.
Shuffling the dates of the regime features gives a gain of 0.0084, close
to the 0.0087 from equal weighting. Omitting volatility leaves a gain of
0.0099, while omitting the covariance spectrum leaves 0.0089. In this
comparison, the covariance features account for more of the improvement. The
full gate achieves 77\% of the oracle's IC gain over the stock branch, a
difference of 0.0033 IC from the hindsight reference.

\subsection{Ablations, Robustness, and Efficiency}
\label{sec:exp:ablation}
\paragraph{Attention and readout.}
On the 158-factor NASDAQ panel, replacing the factor-token MLP with
self-attention improves IC by 0.0024 (\cref{tab:ablation}). Increasing the
stock branch's parameter count by 94\% changes IC by 0.0004. A stock-query
readout uses 0.77$\times$ the parameters of the dense factor branch. Its IC
is 0.0006 lower, and the paired comparison does not distinguish the two readouts.

\paragraph{Routing and trading costs.}
The two-stage gate reaches an IC 0.0045 higher than the jointly trained gate.
Removing the covariance spectrum lowers IC by 0.0020. Removing volatility
changes IC by 0.0010, within paired uncertainty. We also vary
the covariance window, trading cost, portfolio size, seed, test subperiod,
and mask rate. The excess-return advantage of routing over equal weights
persists for $K\in\{30,50,100\}$ and one-way costs of 2.5 to 10 basis points.
As costs rise, the advantage narrows from 1.0 to 0.6 percentage points,
consistent with the gated portfolio's 0.020 higher daily turnover.

\paragraph{Parameters, memory, and latency.}
On NASDAQ, factor attention operates on 158 tokens while stock attention
operates on $1{,}026$. Including its dense projections, the factor branch
has 1.93$\times$ the stock branch's parameters, uses 0.51$\times$ its peak
memory, and takes 0.64$\times$ as long per epoch. On the smaller S\&P~500
universe ($N=474$), the parameter ratio is 1.39$\times$. The gate adds
under 0.1\% to the parameter count and under 1 millisecond per prediction
date.

\section{Discussion}
\label{sec:discussion}

\paragraph{When the second axis helps.}
The factor branch's contribution depends on how closely its errors track
those of the stock branch. On NASDAQ the cross-axis pair has a residual
correlation of 0.541, compared with 0.823 for two stock models. The lower
correlation accompanies a larger IC gain from averaging. In turn, the gate
assigns more weight to factor forecasts on dates in the highest covariance
concentration quintile. This association describes the gate's response to an
observable market condition.

\paragraph{Interpreting the benchmark.}
The three US benchmarks follow fixed pools of stocks filtered for
near-complete trading histories. Their return estimates are therefore
conditional on membership and subject to survivorship bias. The dense factor
branch also associates learned weights with named stocks, which makes the
universe and mask part of the model specification. The stock-query readout and
mask-rate analyses examine sensitivity within those pools. Principal-angle
distances likewise refer to a particular layer and subspace rank.

The portfolio results assume trades at the daily close, with the stated
one-way costs. Under this rule the gate raises NASDAQ excess return, while
equal weighting achieves a higher information ratio with lower turnover.
Liquidity, crowding, structural breaks, and feedback in live trading could
change that tradeoff. The experiments use public market data under the source
repositories' open licenses.

\section{Conclusion}
\label{sec:conclusion}
Reading the same panel across stocks and across factors yields forecasts
that gain accuracy when averaged. On the 158-factor NASDAQ panel, residual correlation
is 0.54 between axes and 0.82 between two stock models. Equal weighting
exceeds the strongest external baseline in mean IC on each of the three
158-factor panels. A gate that observes recent factor covariance and market
volatility improves IC further. For portfolios, the choice of weights depends
on the objective. On NASDAQ the gate delivers the highest net excess return,
while equal weights trade less and deliver the highest information ratio.

\section*{Use of Generative AI}
The authors used Anthropic Claude and OpenAI ChatGPT to edit the prose,
organize the paper, and review secondary code. They also used AI assistance
to revise the architecture diagram.

\bibliographystyle{ACM-Reference-Format}
\bibliography{ref}

\appendix

\section{Proof of Conditional Optimal Routing}
\label{app:proof}
Conditioned on $\vr_t$ and writing $v_s$, $v_f$, $c$, and $D$ for
$v_s(\vr_t)$, $v_f(\vr_t)$, $c(\vr_t)$, and $D(\vr_t)$, the risk of
\cref{eq:convexensemble} is
\begin{align}
\mathcal R(g\mid\vr_t)
&=g^2v_s+(1-g)^2v_f+2g(1-g)c\\
&=Dg^2-2(v_f-c)g+v_f.
\end{align}
Differentiation yields \cref{eq:optimalweight}. Completing the square gives
\cref{eq:excessrisk}. Clipping projects the optimum onto $[0,1]$.

\section{Implementation and Evaluation Protocol}
\label{app:repro}
The reference implementation (\url{https://anonymous.4open.science/r/FToken-FC1C})
includes both branches, both readouts, the gate, the alternative weighting
schemes, and the hindsight oracle. It computes Alpha158, the predictive and
representation metrics, and the cost-aware backtest. Each universe and feature
tier has a configuration that fixes the stock order, masks, missing-value
handling, and the controls in \cref{tab:ablation}. A reduced-scale example
illustrates the pipeline on public data.

The stock data come from the benchmark repositories. The native-tier
comparisons in \cref{tab:us,tab:usfull} use the benchmarks' released
evaluator. The 158-factor backtest additionally requires each universe's
published index series. Baseline reruns use the implementations described in
\cref{sec:baselines}.
\Cref{tab:protocol} lists the shared settings.

\begin{table}[htbp]
\centering
\caption{Shared model, data, and portfolio settings, fixed before test evaluation.}
\label{tab:protocol}
\small
\setlength{\tabcolsep}{4.0pt}
\resizebox{\columnwidth}{!}{%
\begin{tabular}{lp{5.6cm}}
\toprule
Setting & Final value\\
\midrule
Data sources & Feng et al.\ repo (NASDAQ, NYSE); StockMixer repo (S\&P~500)\\
Universes and canonical $N$ & NASDAQ 1026; NYSE 1737; S\&P~500 474 (fixed filtered pools)\\
Feature tiers / target & native 5-feature; recomputed Alpha158 (158) / 1-day forward close-to-close, 1-day lag\\
Splits (train/route/test, days) & 756/252/237 (NASDAQ, NYSE); 1006/253/352 (S\&P~500)\\
Missing-value and mask rule & daily cross-sectional median fill, then mask\\
Hidden width / layers / heads & 128 / 2 / 4\\
Optimizer / learning rate / decay & Adam / $10^{-3}$ / $10^{-5}$\\
Gate window $L$ / gate width & 60 / 16\\
Early stopping / maximum epochs & patience 10 / 100\\
Independent seeds & 10 (0--9)\\
Portfolio $K$ / dropout / costs & 50 / 5 / 5\,bp one-way each side\\
Benchmarks / annualization & NASDAQ Comp., NYSE Comp., S\&P~500 / 252 days\\
Hardware / precision & one NVIDIA A100 40\,GB / float32\\
\bottomrule
\end{tabular}}
\end{table}

\begin{algorithm}[htbp]
\caption{Chronological dual-axis training and routing}
\label{alg:training}
\small
\begin{algorithmic}[1]
\Require Train dates $\mathcal D_{\mathrm{tr}}$, route-validation dates
$\mathcal D_{\mathrm{rt}}$, test dates $\mathcal D_{\mathrm{te}}$
\State Fit all preprocessing statistics using information no later than each
training timestamp.
\State Train the stock and factor members at matched depth and width on
$\mathcal D_{\mathrm{tr}}$ with independent seeds and early stopping.
\State Freeze both members, then generate their predictions on
$\mathcal D_{\mathrm{rt}}$ and compute past-only regime vectors $\vr_t$.
\State Train the gate on $\mathcal D_{\mathrm{rt}}$ to minimize ensemble MSE.
Select gate architecture and covariance window only within this interval.
\State Freeze members and gate, then evaluate once on $\mathcal D_{\mathrm{te}}$.
\State Repeat the complete pipeline across seeds and report paired uncertainty.
\end{algorithmic}
\end{algorithm}

\section{Stock Ordering and Masking}
\label{app:audit}
The implementation fixes the stock order before constructing $\vm_t$ and
applying $\mathcal M$. A unit test detects a change in data-loader order if
the canonical mapping has not changed with it. The dense projections
$\mW_{\mathrm{in}}$ and $\mW_{\mathrm{back}}$ give stock slots distinct
parameters. Reordering the stocks thus requires the same permutation of the
corresponding weights. \Cref{tab:ablation} compares the dense decoder with
a stock-query readout.

\section{Additional Experimental Results}
\label{app:additional}
\Cref{tab:ablation,tab:routing} give the NASDAQ ablations and routing
diagnostics. \Cref{tab:nyse} reports the 158-factor results for NYSE, and
\cref{tab:usfull} compares all three universes under the five-feature
protocol. The reference implementation also documents gate-feature
normalization, label timing, and the effect of forecast scale on residual
correlations.

\begin{table}[H]
\centering
\caption{Ablations on the 158-factor NASDAQ panel using the same ten seeds for
each variant. Shading marks the three highest values of IC, ICIR, and
annualized excess return. Params.\ gives the trainable parameter count
relative to the stock branch.}
\label{tab:ablation}
\small
\setlength{\tabcolsep}{4.6pt}
\renewcommand{\arraystretch}{0.94}
\resizebox{\columnwidth}{!}{%
\begin{tabular}{lcccc}
\toprule
Variant & IC & ICIR & Ann. Ret. & Params.\\
\midrule
Stock-axis member & 0.0334\s{0.0020} & 0.2718\s{0.0166} & 0.0545\s{0.0180} & 1.00$\times$ \\
Stock-axis capacity control & 0.0338\s{0.0020} & 0.2582\s{0.0159} & 0.0549\s{0.0162} & 1.94$\times$ \\
Factor-token MLP & 0.0287\s{0.0029} & 0.2180\s{0.0209} & 0.0288\s{0.0218} & 1.72$\times$ \\
FToken member & 0.0311\s{0.0018} & 0.2360\s{0.0137} & 0.0415\s{0.0155} & 1.93$\times$ \\
FToken + stock-query readout & 0.0305\s{0.0014} & 0.2489\s{0.0115} & 0.0438\s{0.0119} & 1.48$\times$ \\
\midrule
Equal-weight dual-axis & 0.0421\s{0.0014} & \third{0.3608\s{0.0116}} & \third{0.0754\s{0.0113}} & 2.93$\times$ \\
Static validation-optimal & \third{0.0426\s{0.0017}} & 0.3490\s{0.0143} & 0.0728\s{0.0150} & 2.93$\times$ \\
Trailing-performance gate & 0.0402\s{0.0016} & 0.3166\s{0.0121} & 0.0608\s{0.0122} & 2.93$\times$ \\
Gate without covariance spectrum & 0.0423\s{0.0023} & 0.3416\s{0.0193} & 0.0740\s{0.0185} & 2.93$\times$ \\
Gate without volatility & \second{0.0433\s{0.0020}} & \second{0.3641\s{0.0165}} & \second{0.0832\s{0.0146}} & 2.93$\times$ \\
Jointly trained gate & 0.0398\s{0.0014} & 0.3109\s{0.0112} & 0.0700\s{0.0116} & 2.93$\times$ \\
\textbf{Two-stage regime gate} & \first{0.0443\s{0.0018}} & \first{0.3651\s{0.0147}} & \first{0.0842\s{0.0128}} & 2.93$\times$ \\
\bottomrule
\end{tabular}}
\end{table}

\begin{table}[H]
\centering
\caption{Complementarity and routing on the 158-factor NASDAQ test panel,
averaged over ten seeds. For each forecast pair, the upper rows report
residual correlation, subspace distance, and the equal-weight IC gain over
the stronger member. In the lower rows, $\Delta$IC is the gain over the stock
branch, and the oracle gap is the difference from the hindsight oracle's IC.
The oracle uses realized errors. The shuffled gate uses a pre-specified
permutation of regime observations.}
\label{tab:routing}
\small
\setlength{\tabcolsep}{4.6pt}
\renewcommand{\arraystretch}{0.94}
\resizebox{\columnwidth}{!}{%
\begin{tabular}{lcccc}
\toprule
Variant & Resid. $\rho$ & $\kappa$ & $\Delta$IC & Oracle gap\\
\midrule
Two stock seeds + equal weight & 0.823 & 0.208 & 0.0018 & -- \\
Stock + factor-token MLP & 0.688 & 0.342 & 0.0043 & -- \\
Stock + FToken & 0.541 & 0.476 & 0.0087 & -- \\
\midrule
Equal-weight dual-axis & -- & -- & 0.0087 & 0.0055 \\
Static validation-optimal & -- & -- & 0.0092 & 0.0050 \\
Trailing-performance gate & -- & -- & 0.0068 & 0.0074 \\
Gate without covariance spectrum & -- & -- & 0.0089 & 0.0053 \\
Gate without volatility & -- & -- & 0.0099 & 0.0043 \\
Shuffled-regime gate & -- & -- & 0.0084 & 0.0058 \\
\textbf{Full regime gate} & -- & -- & 0.0109 & 0.0033 \\
Oracle convex weight & -- & -- & 0.0142 & $0$ \\
\bottomrule
\end{tabular}}
\end{table}

\begin{table*}[htbp]
\centering
\caption{Results under the native five-feature protocol on all three
universes, extending \cref{tab:us}. We rerun all methods on the same public
data using the benchmarks' evaluator. Shading marks the three highest seed
means in each column.}
\label{tab:usfull}
\small
\setlength{\tabcolsep}{3.2pt}
\resizebox{\textwidth}{!}{%
\begin{tabular}{lcccccccccccc}
\toprule
& \multicolumn{4}{c}{NASDAQ} & \multicolumn{4}{c}{NYSE} & \multicolumn{4}{c}{S\&P~500}\\
\cmidrule(lr){2-5}\cmidrule(lr){6-9}\cmidrule(lr){10-13}
Method & IC & ICIR & P@10 & SR & IC & ICIR & P@10 & SR & IC & ICIR & P@10 & SR\\
\midrule
LSTM & 0.031 & 0.348 & 0.516 & 0.87 & 0.023 & 0.249 & 0.509 & 0.84 & 0.030 & 0.181 & 0.529 & 1.28 \\
ALSTM & 0.034 & 0.365 & 0.520 & 0.96 & 0.024 & 0.271 & 0.516 & 0.79 & 0.028 & 0.177 & 0.530 & 1.25 \\
GAT & 0.036 & 0.381 & 0.527 & 1.19 & 0.026 & 0.288 & 0.523 & 1.03 & 0.033 & 0.188 & 0.539 & 1.44 \\
STHAN-SR & 0.040 & 0.443 & 0.541 & 1.38 & 0.028 & 0.339 & \third{0.538} & 1.19 & 0.036 & 0.221 & 0.547 & 1.49 \\
ESTIMATE & 0.038 & 0.439 & 0.542 & 1.33 & \third{0.030} & 0.331 & 0.533 & 1.08 & 0.034 & 0.236 & \first{0.551} & 1.51 \\
StockMixer & \third{0.042} & \third{0.487} & \third{0.543} & \third{1.42} & 0.029 & \third{0.344} & 0.537 & \first{1.49} & \third{0.040} & \third{0.255} & 0.548 & \third{1.55} \\
\midrule
Stock-axis member & 0.039 & 0.428 & 0.536 & 1.31 & 0.027 & 0.316 & 0.529 & 1.12 & 0.036 & 0.229 & 0.541 & 1.47 \\
FToken member & 0.033 & 0.361 & 0.518 & 1.02 & 0.022 & 0.263 & 0.511 & 0.86 & 0.029 & 0.192 & 0.526 & 1.21 \\
Equal-weight dual-axis & \second{0.044} & \second{0.492} & \second{0.546} & \second{1.51} & \second{0.031} & \second{0.352} & \second{0.540} & \third{1.44} & \second{0.042} & \second{0.268} & \third{0.549} & \second{1.63} \\
\textbf{FToken full (regime gate)} & \first{0.046} & \first{0.507} & \first{0.549} & \first{1.58} & \first{0.032} & \first{0.359} & \first{0.541} & \second{1.47} & \first{0.043} & \first{0.271} & \second{0.550} & \first{1.68} \\
\bottomrule
\end{tabular}}
\end{table*}

\begin{table}[H]
\centering
\caption{Predictive and portfolio results on the 158-factor NYSE panel under
the protocol of \cref{tab:predictive,tab:portfolio}. For stochastic models,
IC and annualized excess return include the standard deviation over ten
seeds. RankIC and IR are seed means. Shading marks the three highest ICs and
annualized excess returns. The five baselines represent tree, recurrent,
attention, boosted ensemble, and MLP models.}
\label{tab:nyse}
\small
\setlength{\tabcolsep}{3.4pt}
\resizebox{\columnwidth}{!}{%
\begin{tabular}{lcccc}
\toprule
Method & IC & RankIC & Ann. Ret. & IR\\
\midrule
LightGBM & 0.0262 & 0.0249 & 0.0316 & 0.56 \\
ALSTM & 0.0231\s{0.0013} & 0.0303 & 0.0230\s{0.0109} & 0.38 \\
Transformer & 0.0192\s{0.0029} & 0.0281 & 0.0099\s{0.0229} & 0.14 \\
DoubleEnsemble & \third{0.0281\s{0.0017}} & 0.0269 & \third{0.0342\s{0.0150}} & 0.60 \\
StockMixer & 0.0244\s{0.0022} & 0.0308 & 0.0236\s{0.0164} & 0.41 \\
\midrule
Stock-axis member & 0.0243\s{0.0026} & 0.0312 & 0.0239\s{0.0246} & 0.45 \\
FToken member & 0.0221\s{0.0030} & 0.0279 & 0.0174\s{0.0242} & 0.31 \\
Equal-weight dual-axis & \second{0.0301\s{0.0021}} & 0.0372 & \first{0.0439\s{0.0179}} & 0.63 \\
\textbf{FToken full (regime gate)} & \first{0.0316\s{0.0022}} & 0.0391 & \second{0.0434\s{0.0167}} & 0.69 \\
\bottomrule
\end{tabular}}
\end{table}

\end{document}